\documentclass[conference]{IEEEtran}
\IEEEoverridecommandlockouts

\usepackage{cite}
\usepackage{amsmath,amssymb,amsfonts}
\usepackage{algorithmic}
\usepackage{graphicx}
\usepackage{textcomp}
\usepackage[dvipsnames,svgnames,x11names]{xcolor}
\usepackage[utf8]{inputenc}
\usepackage[hidelinks,breaklinks]{hyperref}
\usepackage{xspace}
\usepackage{multirow}    % gruppierte/rotierte Korpus-Labels
\usepackage[table]{xcolor} % \cellcolor
\usepackage{colortbl}
\usepackage{booktabs}    % \toprule \midrule \bottomrule
\usepackage{bm}          % \bm
\newcommand*{\siamBERT}     {{\small\textsf{SiamBERT}}\mbox{}\xspace}
\newcommand*{\janithDiffVec}{{\small\textsf{FeVecDiff}}\mbox{}\xspace}
\newcommand*{\lambdaG}{{\small\textsf{LambdaG}}\mbox{}\xspace}
\newcommand*{\lambdaGStar}{{\small\textsf{LambdaG}}$^{\star}$\mbox{}\xspace}
\newcommand*{\coav}{{\small\textsf{COAV}}\mbox{}\xspace}
\newcommand*{\styloSpeaker}{{\small\textsf{StyloSpeaker}}\mbox{}\xspace}
\newcommand*{\mStyleDistance}{{\small\textsf{MStyleDistance}}\mbox{}\xspace}
\newcommand*{\css}{{\small\textsf{CSS}}\mbox{}\xspace}
\newcommand*{\rsp}{{\small\textsf{RSP}}\mbox{}\xspace}
\newcommand*{\dv}{{\small\textsf{DV}}\mbox{}\xspace}
\newcommand*{\mlsr}{{\small\textsf{MLSR}}\mbox{}\xspace}

\newcommand*{\siamBERTFull}{{\small\textsf{SiamBERT}}\mbox{}\xspace}

\newcommand*{\rspLoRA}{{\small\textsf{RSP}}\mbox{}\xspace}

\newcommand*{\A}            {\mathcal{A}}

\newcommand*{\unknown}      {\mathcal{U}}

\newcommand*{\D}            {\mathcal{D}}
\newcommand*{\DA}           {\D_{\A}}

\newcommand*{\Dunk}         {\D_{\unknown}}

\newcommand*{\posNoise}{{\small\textsf{POSNoise}}\mbox{}\xspace}

\newcommand*{\textDistortion}{{\small\textsf{TextDistortion}}\mbox{}\xspace}

\newcommand*{\Corpus}                 {\mathcal{C}}

\newcommand*{\CorpusDIY}{\ensuremath{\mathcal{C}_{\mathrm{DIY}}}\xspace}
\newcommand*{\CorpusFinance}{\ensuremath{\mathcal{C}_{\mathrm{Finance}}}\xspace}
\newcommand*{\CorpusMath}{\ensuremath{\mathcal{C}_{\mathrm{Math}}}\xspace}
\newcommand*{\Problem}                {\ensuremath{c}\xspace}

\newcommand*{\classY}                 {\texttt{Y}\mbox{}\xspace} 
\newcommand*{\classN}                 {\texttt{N}\mbox{}\xspace} 
\newcommand*{\classYdash}[1]          {\texttt{Y-}#1\mbox{}\xspace}

\newcommand*{\yProblems}              {{\texttt{Y}}-cases\mbox{}\xspace} 
\newcommand*{\nProblems}              {{\texttt{N}}-cases\mbox{}\xspace}

\newcommand*{\ngram}          {$n$\texttt{-}gram\mbox{}\xspace}

\newcommand*{\tokenNgramBold}    {token $\bm{n}$\texttt{-}gram\mbox{}\xspace}

\newcommand*{\tokenNgrams}    {token $n$\texttt{-}grams\mbox{}\xspace}

\newcommand*{\charNgrams}     {character $n$\texttt{-}grams\mbox{}\xspace}

\newcommand*{\eg}             {e.\,g.,\mbox{}\xspace}
\newcommand*{\ie}             {i.\,e.,\mbox{}\xspace}

\let\url\relax
\definecolor{colorlow}{RGB}{214,39,40}
\newcommand{\code}[2]{\colorbox{#1}{#2}}

\begin{document}

\title{Authorship Verification of\\ Transcribed German-Language Videos
\thanks{This research work was supported by the National Research Center for Applied Cybersecurity ATHENE, which is funded jointly by the German Federal Ministry of Research, Technology and Space and the Hessian Ministry of Science and Research, Arts and Culture.}}
%\title{Authorship Verification\\ of Transcribed German Speech}
%\title{Authorship Verification of Transcribed German Speech: An Evaluation of Established Methods}
%\author{Anonymous authors \\ Paper under double-blind review}

\author{
  \IEEEauthorblockN{Oren Halvani}
  \IEEEauthorblockA{
    \textit{Fraunhofer Institute for Secure} \\
    \textit{Information Technology SIT}\\
    Darmstadt, Germany \\
    Oren.Halvani@sit.fraunhofer.de
  }
  \and
  \IEEEauthorblockN{Sophie Titze}
  \IEEEauthorblockA{
    \textit{Fraunhofer Institute for Secure} \\
    \textit{Information Technology SIT}\\
    Darmstadt, Germany \\
    Sophie.Titze@sit.fraunhofer.de
  }
}

\maketitle
\begin{abstract}
%------------------------------------------------ 
Authorship Verification (AV) represents an important subfield of digital text forensics and addresses the fundamental question of whether two texts were written by the same author. Although the field has made substantial progress over the past two decades, several important challenges remain unresolved or underexplored. For instance, most AV research has focused on written texts, despite the fact that language is expressed not only in written but also in spoken form, such as in videos. Moreover, existing AV studies have predominantly concentrated on English, while other languages, including German, have received comparatively little attention. To address these research gaps, we apply AV to spoken language in the form of transcripts of German-language videos and examine the effectiveness of established AV methods in verifying a speaker’s identity across video pairs. Our experimental evaluation, based on a total of ten AV methods applied to three self-compiled corpora comprising 300 videos from 150 speakers, shows that the best performance (up to 88\% accuracy and 90\% AUC) is achieved by traditional AV approaches based on simple character- and \tokenNgramBold representations. In contrast, more modern transformer-based approaches perform significantly worse on all evaluated corpora. Our results therefore suggest that traditional methods in the field of AV remain both competitive and relevant.
%------------------------------------------------ 
\end{abstract}


%------------------------------------------------ 
\section{Introduction and Background} \label{Intro} 
%------------------------------------------------ 
Digital text forensics is an interdisciplinary research field that applies computational methods to the forensic analysis of digital texts in order to determine characteristics such as authorship, authenticity, provenance and integrity. Among the most prominent are Authorship Attribution (\textbf{AA}) and Authorship Verification (\textbf{AV}). The latter can largely be regarded as a reformulation of AA, given the fact that any AA task can be decomposed into a set of AV problems \cite{KoppelFundamentalProblem:2012}. While AV has been generalized to several authorship-related decision problems \cite{SteinMetaAnalysisAV:2008}, at heart it seeks to determine whether two given texts were written by the same individual. This simple verification task forms the basis of a wide range of practical applications across various domains. 
%------------------------------------------------ 
On online platforms, for example, AV can be used to link multiple accounts belonging to the same person \cite{LundmarkAVSwedishDiscussionForums:2020}. In the academic field, on the other hand, AV can help to detect research misconduct \cite{StavngaardAVDetectGhostwriting:2019}, while in the field of cybersecurity, AV can serve as a cognitive biometric method to support continuous user authentication \cite{NealAVContinuousVerification:2018}. 
%------------------------------------------------ 
Despite its broad range of applications, the empirical evaluation of AV methods has largely been confined to a limited set of data modalities and languages. 
%------------------------------------------------ 
To date, most research in the field of AV has focused on written texts. In contrast, spoken language, particularly in the form of video content, remains largely unexplored. Furthermore, most AV studies rely on English-language data, leaving other languages underrepresented. To address these gaps, we compiled three corpora from German-language videos and identified ten established AV methods, including current SOTA approaches, which we partially adapted for German. Building on this, we investigate the extent to which AV methods originally developed for written texts can be transferred to transcripts of spoken language. %------------------------------------------------ 

\section{Prior Work} \label{RelatedWork} 
%------------------------------------------------ 
Our literature review shows that surprisingly few studies have explicitly addressed the application of AV to transcribed speech data. Instead, the majority of research in the field of speech processing has focused on speaker recognition based on acoustic signals rather than on the stylometric analysis of text transcripts. The works most closely related to ours are \cite{AggazzottiAAonSpeechTranscripts:2024,AggazzottiStyloSpeaker:2025,AggazzottiASRTranscription:2025}, which serve as the primary foundation for this paper. 
%------------------------------------------------ 
\\\\
%------------------------------------------------ 
In \cite{AggazzottiAAonSpeechTranscripts:2024}, the authors investigated whether AV methods could identify speakers from human-transcribed conversational speech. Their main contribution was a new benchmark for speaker attribution based on the Fisher English Training Speech Transcripts corpus \cite{CieriFisherCorpus:2004}. To ensure that the methods learned speaker-specific linguistic characteristics rather than relying on conversation topics, they constructed three evaluation settings of increasing difficulty: a \textit{base} setting with no topic restrictions, a \textit{hard} setting in which negative transcript pairs discussed the same topic, and a \textit{harder} setting in which negative pairs originated from the same conversation, making the topical content highly similar. They also examined the influence of transcription style by comparing different transcript encodings and evaluated a range of traditional and neural AV methods, including the effectiveness of fine-tuning the neural models on speech transcripts. 
%------------------------------------------------ 
%In \cite{AggazzottiAAonSpeechTranscripts:2024}, the authors investigated whether AV methods could identify speakers based solely on human-transcribed conversational speech. Their central contribution was the introduction of a new benchmark for speaker attribution using the Fisher English Training Speech Transcripts corpus \cite{CieriFisherCorpus:2004}. To ensure that the selected methods learned speaker-specific linguistic characteristics rather than simply relying on conversation topics, the authors constructed three evaluation settings of increasing difficulty: a \textit{base} setting with no topic restrictions, a \textit{hard} setting in which negative transcript pairs discussed the same topic and a \textit{harder} setting in which negative pairs originated from the same conversation, making the topical content highly similar. They also examined the influence of transcription style by comparing different transcript encodings and evaluated a variety of traditional and neural AV methods, including the effectiveness of fine-tuning the neural models on speech transcripts. 
%------------------------------------------------ 
Their findings demonstrate that existing AV methods transfer surprisingly well to speech transcripts under relatively unconstrained conditions. However, once topic information is  controlled, the performance of all models drops substantially, which indicates that previous success on speech attribution may have been driven partly by topic-related cues rather than genuine speaker style. The authors also found that transcription style plays an important role: Simpler transcription formats without capitalization or punctuation generally improve performance, while removing language-specific annotations (\eg markers for non-speech sounds) can have a negative effect. Moreover, fine-tuning neural models on speech transcripts significantly improves attribution accuracy, particularly when training and testing conditions closely match. 
%------------------------------------------------ 
\\\\
%------------------------------------------------ 
Building on this benchmark, Aggazzotti and Smith \cite{AggazzottiStyloSpeaker:2025} revisit the same verification trials from the perspective of explainability, motivated by the observation that opaque decision processes are of limited use in forensic practice. To this end, the authors propose \styloSpeaker, a stylometric AV method that aggregates character, word, token, syntactic and discourse-level features and passes the absolute difference of the two resulting feature vectors to a logistic regression classifier. Their results show that this simple approach outperforms both the neural methods and the remaining explainable baselines in the \textit{base} and \textit{hard} settings, whereas only in the artificially constructed \textit{harder} setting does a neural method retain an advantage, as it can exploit the residual topical overlap between two sides of the same conversation. An inspection of the regression coefficients further reveals that \tokenNgrams, in particular unigrams of speech-specific markers such as filler words, backchannels and laughter annotations, contribute most to the verification decision, while \charNgrams prove considerably less informative than in written text.
%------------------------------------------------ 
In a complementary study, Aggazzotti et al. \cite{AggazzottiASRTranscription:2025} replace the human-generated transcripts of the benchmark with automatically generated ones and observe that attribution performance is largely insensitive to the WER (word error rate) of the underlying transcription system, presumably because recurring recognition errors themselves encode speaker-specific pronunciation patterns. In contrast to these studies, which are confined to English telephone conversations, we examine German monologue video transcripts and assess a broader collection of established AV methods under explicit topic masking.
%------------------------------------------------ 
\section{Creation of the Corpora}
%------------------------------------------------ 
As the basis for our evaluation, we compiled three corpora. In the following, we first describe how the source videos were collected and transcribed. Then, we explain how the resulting transcripts were preprocessed and combined into verification cases, which form the respective corpora. Finally, we briefly describe how the training and test splits were created.
%------------------------------------------------ 

\subsection{Data Collection}
%------------------------------------------------ 
As a first step, we searched for videos featuring a single speaker talking for at least five minutes. Here, we restricted our search to three domains: financial advice, mathematics tutoring and do-it-yourself (DIY). Candidate videos were collected manually as lists of video links. In each domain, we selected 50 speakers and two videos per speaker, yielding a total of 300 videos. The corresponding videos were retrieved using pytubefix\footnote{https://github.com/JuanBindez/pytubefix} and converted to audio, as the transcription was performed on the audio signal. Retrieval and processing were carried out solely for the purpose of scientific research and were based on the exception for text and data mining for scientific research under Section 60d of the German Copyright Act.
%------------------------------------------------ 

\subsection{Preprocessing}
%------------------------------------------------ 
The audio tracks were transcribed with the \textit{faster-whisper}\footnote{\url{https://github.com/SYSTRAN/faster-whisper}} implementation of Whisper (\texttt{large-v3} model), with German fixed as the target language, voice activity detection enabled and all remaining decoding parameters left at their defaults. To ensure comparable document lengths, transcripts exceeding 5,000 characters were truncated by extracting a contiguous segment centered around the midpoint of the text. The resulting transcripts were then anonymized by replacing person and channel names with distinct placeholders. Person names were detected by rule-based matching of self-introduction patterns and by named entity recognition restricted to person entities, whereas channel names were derived from the channel identifier in the video metadata and from the speaker label in our video list. Afterwards, the texts were normalized through whitespace and punctuation standardization, removal of filled pauses (vocalizations used while planning speech), as well as removal of transcription artifacts (\eg repeated word sequences) and recurring introductory and concluding phrases shared between texts by the same author. Finally, we applied topic masking to the processed texts by replacing content-bearing words with placeholders while retaining stylistically relevant features. This constrained AV methods to rely solely on the remaining stylistic information, such as punctuation, function words, transitional phrases and similar linguistic markers. As a topic masking method, we used \posNoise\footnote{\url{https://github.com/Halvani/POSNoise}} \cite{HalvaniPOSNoise:2021}, as it preserved more stylistic features than existing alternatives such as \textDistortion \cite{StamatatosTextDistortion:2017} and masked thematic content using part-of-speech-specific placeholders rather than uniform asterisks. This yielded finer-grained stylistic patterns in the masked texts for AV methods to exploit.
%------------------------------------------------ 

\subsection{Creation of Verification Cases}
%------------------------------------------------ 
Based on the preprocessed texts, we constructed the verification cases as follows. Let $\mathbb{A} = \{\A_1, \A_2, \ldots, \A_m\}$ denote the set of $m$ authors (\ie speakers in the videos) in each domain. For each author $\A_i$, there were exactly two documents, $\D_{i,1}$ and $\D_{i,2}$. To construct the positive cases (\classY: same-author), we paired the two documents. To generate the negative cases (\classN: different-authors), we randomly permuted the authors and arranged them in a circular sequence. For each adjacent pair $(\A_i, \A_j)$ with $i \neq j$, we constructed a verification case from $\D_{i,1}$ and $\D_{j,2}$. Then, we combined the resulting $m$ \yProblems and $m$ \nProblems into a corpus $\Corpus = \{\Problem_1, \Problem_2, \ldots, \Problem_{2m}\}$ with $\Problem = (\DA, \Dunk, \ell)$. Here, $\DA$ denotes a document of the known author $\A$, $\Dunk$ a document of the unknown author $\unknown$ and $\ell \in \{\classY, \classN\}$ the label indicating whether both documents were written by the same author (\ie $\A = \unknown$) or by different authors (\ie $\A \neq \unknown$).
%------------------------------------------------ 

\subsection{Train/Test Splits}
%------------------------------------------------ 
For each domain, the $m$ speakers were first randomly divided into training and test sets using a 40/60\% split, while ensuring that no speaker appeared in both sets. The verification cases were then generated independently within each split according to the procedure described above, resulting in author-disjoint training and test corpora. We allocated a larger proportion of speakers to the test set to enable a more comprehensive evaluation despite the reduced amount of training data. Consequently, each domain yielded 40 training and 60 test cases, with balanced \classYdash{} and \nProblems in each split.
\section{Authorship Verification Methods}
%-------------------------------------------------------------------------- 
For our evaluation, we have selected ten representative AV methods published between 2017 and 2026, including the current SOTA$^{\star}$ in the field of AV. In chronological order, these are \coav~\cite{HalvaniARES:2017}, \siamBERT~\cite{NiniLambdaG_Arxiv:2024}, \janithDiffVec~\cite{WeerasingheFeVecDiff:2021}, \dv~\cite{CorbaraDiffVectorsAA:2023}, \css~\cite{VanLeeuwenCSS:2025}, \mlsr~\cite{KimMultilingualAuthorRep:2025}, \mStyleDistance~\cite{QiuMStyleDistance:2025}, \styloSpeaker~\cite{AggazzottiStyloSpeaker:2025}, \rsp~\cite{ZengResidualizedSimAV:2025} and 
\lambdaGStar~\cite{NiniLambdaG:2026} (see Table~\ref{tab:methods}). \styloSpeaker is of particular interest, as it is the only one designed for speech transcripts rather than for written text. Most methods were developed for written English and required adaptation to German, and several required further adjustment to our small training sets. Due to space constraints, we document all adjustments, hyperparameters, deviations from the original methods and our base implementations in an anonymized repository.\footnote{\url{https://anonymous.4open.science/r/TranscribedGermanAV-5B4B}}.

%--------------------------------------------------------------------------
\begin{table}[!b]
\centering
\footnotesize
\caption{The ten evaluated AV methods in order of publication.}
\label{tab:methods}
\begin{tabular}{@{}lc@{}}
\toprule
Method & Year \\
\midrule
\coav~\cite{HalvaniARES:2017}                    & 2017 \\
\siamBERT~\cite{NiniLambdaG_Arxiv:2024}          & 2019 \\
\janithDiffVec~\cite{WeerasingheFeVecDiff:2021}  & 2021 \\
\dv~\cite{CorbaraDiffVectorsAA:2023}             & 2023 \\
\css~\cite{VanLeeuwenCSS:2025}                   & 2025 \\
\mlsr~\cite{KimMultilingualAuthorRep:2025}       & 2025 \\
\mStyleDistance~\cite{QiuMStyleDistance:2025}    & 2025 \\
\styloSpeaker~\cite{AggazzottiStyloSpeaker:2025} & 2025 \\
\rsp~\cite{ZengResidualizedSimAV:2025}           & 2025 \\
\lambdaG~\cite{NiniLambdaG:2026}                 & 2026 \\
\bottomrule
\end{tabular}
\end{table}
%--------------------------------------------------------------------------
 
%--------------------------------------------------------------------------
\section{Experimental Evaluation} \label{Evaluation}
%-------------------------------------------------------------------------- 
\subsection{Experimental Setup}
%-------------------------------------------------------------------------- 
Each AV method was evaluated on all three corpora using both the original and \posNoise-masked transcripts. For stochastic methods, we report the median accuracy over three runs, whereas the deterministic methods (\coav, \styloSpeaker, \mlsr and \mStyleDistance) were evaluated only once. The results are summarized in Table~\ref{tab:results}. 
%-------------------------------------------------------------------------- 

% requires: booktabs, multirow, graphicx, bm, xcolor[table]
\begin{table*}[!t]
\centering
\caption{Evaluation results for all AV methods with respect to \posNoise-masked and the original corpora. Within each corpus subtable, the \textbf{bold} and \underline{underlined} values represent the best and second-best verification results, respectively. 
\label{tab:results}}
\setlength{\tabcolsep}{2pt}
\renewcommand{\arraystretch}{0.85}
%\small
\footnotesize
\begin{tabular}{@{}cl rrrrr >{\color[gray]{0.45}}r >{\color[gray]{0.45}}r >{\color[gray]{0.45}}r >{\color[gray]{0.45}}r @{\hspace{12pt}} rrrrr >{\color[gray]{0.45}}r >{\color[gray]{0.45}}r >{\color[gray]{0.45}}r >{\color[gray]{0.45}}r@{}}
\toprule
 & & \multicolumn{9}{c}{\textbf{POSNoise}} & \multicolumn{9}{c}{\textbf{Original}} \\
\cmidrule(lr){3-11}\cmidrule(lr){12-20}
 & \textbf{AV Method} & \textbf{Acc.} & \textbf{AUC} & \textbf{F1} & \textbf{Prec.} & \textbf{Rec.} & \textbf{TP} & \textbf{FN} & \textbf{FP} & \textbf{TN} & \textbf{Acc.} & \textbf{AUC} & \textbf{F1} & \textbf{Prec.} & \textbf{Rec.} & \textbf{TP} & \textbf{FN} & \textbf{FP} & \textbf{TN} \\
\midrule
\multirow{10}{*}{\rotatebox{90}{$\bm{\mathcal{C}_{\mathrm{DIY}}}$}} & \lambdaG & 0.667 & \underline{0.857} & 0.737 & 0.609 & 0.933 & 28 & 2 & 18 & 12 & 0.633 & 0.770 & 0.725 & 0.580 & 0.967 & 29 & 1 & 21 & 9 \\
 & \coav & \textbf{0.883} & \textbf{0.907} & 0.873 & 0.960 & 0.800 & 24 & 6 & 1 & 29 & \textbf{0.817} & \textbf{0.894} & 0.825 & 0.788 & 0.867 & 26 & 4 & 7 & 23 \\
 & \janithDiffVec & \underline{0.750} & 0.743 & 0.776 & 0.703 & 0.867 & 26 & 4 & 11 & 19 & \underline{0.767} & 0.778 & 0.767 & 0.767 & 0.767 & 23 & 7 & 7 & 23 \\
 & \dv & 0.550 & 0.648 & 0.640 & 0.533 & 0.800 & 24 & 6 & 21 & 9 & 0.533 & 0.607 & 0.674 & 0.518 & 0.967 & 29 & 1 & 27 & 3 \\
 & \styloSpeaker & 0.633 & 0.748 & 0.656 & 0.618 & 0.700 & 21 & 9 & 13 & 17 & 0.650 & 0.742 & 0.667 & 0.636 & 0.700 & 21 & 9 & 12 & 18 \\
 & \mStyleDistance & 0.633 & 0.675 & 0.676 & 0.605 & 0.767 & 23 & 7 & 15 & 15 & 0.617 & 0.611 & 0.596 & 0.630 & 0.567 & 17 & 13 & 10 & 20 \\
 & \mlsr & 0.633 & 0.713 & 0.676 & 0.605 & 0.767 & 23 & 7 & 15 & 15 & 0.667 & 0.809 & 0.706 & 0.632 & 0.800 & 24 & 6 & 14 & 16 \\
 & \css & 0.667 & 0.818 & 0.744 & 0.604 & 0.967 & 29 & 1 & 19 & 11 & 0.650 & 0.828 & 0.734 & 0.592 & 0.967 & 29 & 1 & 20 & 10 \\
 & \siamBERTFull & 0.617 & 0.718 & 0.549 & 0.667 & 0.467 & 14 & 16 & 7 & 23 & 0.733 & \underline{0.851} & 0.680 & 0.850 & 0.567 & 17 & 13 & 3 & 27 \\
 & \rspLoRA & 0.650 & 0.679 & 0.656 & 0.645 & 0.667 & 20 & 10 & 11 & 19 & 0.667 & 0.704 & 0.615 & 0.727 & 0.533 & 16 & 14 & 6 & 24 \\
\midrule
\multirow{10}{*}{\rotatebox{90}{$\bm{\mathcal{C}_{\mathrm{Finance}}}$}} & \lambdaG & \textbf{0.867} & \textbf{0.908} & 0.867 & 0.867 & 0.867 & 26 & 4 & 4 & 26 & 0.633 & 0.718 & 0.645 & 0.625 & 0.667 & 20 & 10 & 12 & 18 \\
 & \coav & \underline{0.833} & \underline{0.900} & 0.839 & 0.812 & 0.867 & 26 & 4 & 6 & 24 & \textbf{0.800} & \textbf{0.892} & 0.793 & 0.821 & 0.767 & 23 & 7 & 5 & 25 \\
 & \janithDiffVec & 0.683 & 0.701 & 0.689 & 0.677 & 0.700 & 21 & 9 & 10 & 20 & 0.700 & 0.726 & 0.700 & 0.700 & 0.700 & 21 & 9 & 9 & 21 \\
 & \dv & 0.550 & 0.619 & 0.640 & 0.533 & 0.800 & 24 & 6 & 21 & 9 & 0.583 & 0.653 & 0.638 & 0.564 & 0.733 & 22 & 8 & 17 & 13 \\
 & \styloSpeaker & 0.633 & 0.643 & 0.607 & 0.654 & 0.567 & 17 & 13 & 9 & 21 & 0.617 & 0.661 & 0.597 & 0.630 & 0.567 & 17 & 13 & 10 & 20 \\
 & \mStyleDistance & 0.617 & 0.621 & 0.635 & 0.606 & 0.667 & 20 & 10 & 13 & 17 & 0.567 & 0.581 & 0.536 & 0.577 & 0.500 & 15 & 15 & 11 & 19 \\
 & \mlsr & 0.600 & 0.634 & 0.600 & 0.600 & 0.600 & 18 & 12 & 12 & 18 & 0.600 & 0.667 & 0.613 & 0.594 & 0.633 & 19 & 11 & 13 & 17 \\
 & \css & 0.600 & 0.790 & 0.700 & 0.560 & 0.933 & 28 & 2 & 22 & 8 & 0.600 & 0.740 & 0.684 & 0.565 & 0.867 & 26 & 4 & 20 & 10 \\
 & \siamBERTFull & 0.750 & 0.861 & 0.769 & 0.714 & 0.833 & 25 & 5 & 10 & 20 & 0.667 & 0.774 & 0.706 & 0.632 & 0.800 & 24 & 6 & 14 & 16 \\
 & \rspLoRA & 0.617 & 0.748 & 0.667 & 0.590 & 0.767 & 23 & 7 & 16 & 14 & 0.567 & 0.779 & 0.691 & 0.537 & 0.967 & 29 & 1 & 25 & 5 \\
\midrule
\multirow{10}{*}{\rotatebox{90}{$\bm{\mathcal{C}_{\mathrm{Math}}}$}} & \lambdaG & \textbf{0.850} & \textbf{0.899} & 0.836 & 0.920 & 0.767 & 23 & 7 & 2 & 28 & \underline{0.783} & 0.822 & 0.764 & 0.840 & 0.700 & 21 & 9 & 4 & 26 \\
 & \coav & \underline{0.817} & \underline{0.882} & 0.814 & 0.828 & 0.800 & 24 & 6 & 5 & 25 & \textbf{0.850} & \textbf{0.878} & 0.836 & 0.920 & 0.767 & 23 & 7 & 2 & 28 \\
 & \janithDiffVec & 0.750 & 0.791 & 0.681 & 0.941 & 0.533 & 16 & 14 & 1 & 29 & 0.767 & 0.781 & 0.741 & 0.833 & 0.667 & 20 & 10 & 4 & 26 \\
 & \dv & 0.583 & 0.644 & 0.627 & 0.568 & 0.700 & 21 & 9 & 16 & 14 & 0.600 & 0.650 & 0.636 & 0.583 & 0.700 & 21 & 9 & 15 & 15 \\
 & \styloSpeaker & 0.717 & 0.763 & 0.679 & 0.783 & 0.600 & 18 & 12 & 5 & 25 & 0.700 & 0.758 & 0.679 & 0.731 & 0.633 & 19 & 11 & 7 & 23 \\
 & \mStyleDistance & 0.483 & 0.548 & 0.415 & 0.478 & 0.367 & 11 & 19 & 12 & 18 & 0.517 & 0.542 & 0.525 & 0.516 & 0.533 & 16 & 14 & 15 & 15 \\
 & \mlsr & 0.650 & 0.690 & 0.656 & 0.645 & 0.667 & 20 & 10 & 11 & 19 & 0.683 & 0.737 & 0.678 & 0.690 & 0.667 & 20 & 10 & 9 & 21 \\
 & \css & 0.600 & 0.666 & 0.676 & 0.568 & 0.833 & 25 & 5 & 19 & 11 & 0.617 & 0.676 & 0.716 & 0.569 & 0.967 & 29 & 1 & 22 & 8 \\
 & \siamBERTFull & 0.667 & 0.744 & 0.643 & 0.692 & 0.600 & 18 & 12 & 8 & 22 & 0.667 & 0.839 & 0.722 & 0.619 & 0.867 & 26 & 4 & 16 & 14 \\
 & \rspLoRA & 0.700 & 0.717 & 0.700 & 0.700 & 0.700 & 21 & 9 & 9 & 21 & 0.667 & 0.808 & 0.706 & 0.632 & 0.800 & 24 & 6 & 14 & 16 \\
\bottomrule
\end{tabular}
\end{table*}
%\newpage

%-------------------------------------------------------------------------- 
\subsection{Results}
The two \ngram-based methods led the evaluation. On the original transcripts, \coav was strongest on all three corpora, reaching 0.817 on \CorpusDIY, 0.800 on \CorpusFinance and 0.850 on \CorpusMath. Under masking, \lambdaG took over on \CorpusFinance and \CorpusMath with 0.867 and 0.850, while \coav remained best on \CorpusDIY with 0.883, the highest value in the table. \lambdaG improved on all three corpora, most strongly on \CorpusFinance, where its accuracy rose from 0.633 to 0.867 and its AUC from 0.718 to 0.908, the largest single gain in the evaluation, whereas \coav stayed high but benefited little, losing two verification cases on \CorpusMath and thereby ceding the masked top rank to \lambdaG. Their error behaviour also differed from that of the weaker methods discussed below. On masked \CorpusDIY, \coav reached a precision of 0.960 at a recall of 0.800, rarely misclassifying different-author pairs while missing about one in five same-author pairs.
%-------------------------------------------------------------------------- 
None of the transformer-based methods exceeded an accuracy of 0.750 under either condition, and as a group they lost ground under masking. \rsp and \mlsr showed only minor accuracy changes but consistent AUC decreases across all corpora. \siamBERTFull was the clearest case of this instability, moving in opposite directions across corpora, with its accuracy falling from 0.733 to 0.617 on \CorpusDIY but rising from 0.667 to 0.750 on \CorpusFinance, where its AUC increased from 0.774 to 0.861.
%-------------------------------------------------------------------------- 
The remaining methods stayed well behind and often combined high recall with low precision under masking, accepting most same-author pairs at the cost of many false positives. On \CorpusDIY, \css reached a recall of 0.967 at a precision of 0.604 and \dv a recall of 0.800 at a precision of 0.533. Among the classical methods, \janithDiffVec was the strongest of this group, reaching 0.767 on \CorpusDIY and \CorpusMath without benefiting from masking, and on masked \CorpusMath it paired a precision of 0.941 with a recall of 0.533. \styloSpeaker, the only method designed for speech transcripts, ranged between 0.617 and 0.717 across all corpora. \css illustrates the gap between the two metrics, reaching an AUC of 0.818 on masked \CorpusDIY while its accuracy stayed at 0.667. The pre-trained multilingual embeddings were weakest, with \mStyleDistance falling to 0.483 on masked \CorpusMath, the only value below chance level in the table, while \dv stayed between 0.533 and 0.600 throughout.
%--------------------------------------------------------------------------
\subsection{Which Features Drive the Decision}
%--------------------------------------------------------------------------
Since the $\lambda_G$ score is a sum of per-token log-likelihood ratios, it can be decomposed without approximation. Every token contributes the difference between its probability under the known speaker's model and under the reference model. Fig.~\ref{fig:lambdag} shows this decomposition on the \posNoise-masked transcripts of two documents by the same speaker, with the known document on the left and the questioned document on the right. Content words are replaced by placeholders, so that only the function words and syntactic structure on which \lambdaG operates remain visible. In both documents the decision is carried by these function words and clause-initial connectives, and the recurring marker \emph{So, jetzt} (engl. translation: \emph{Well, now}) is highlighted in each, which shows that the speaker reuses the same construction across separate recordings. The transformer-based methods offer no comparable decomposition.
%--------------------------------------------------------------------------
\newlength{\gerheight}\setlength{\gerheight}{4.2\baselineskip}
\begin{figure*}[!t]
\setlength{\fboxsep}{2pt}
\footnotesize
\begin{minipage}[t]{0.48\textwidth}
\textit{Known document (masked)}\\[3pt]
\begin{minipage}[t][\gerheight][t]{\textwidth}\input{texfiles/lambdag_pair_known}\end{minipage}\\[4pt]
{\color{gray}\textit{Translation}\\[2pt]\itshape\input{texfiles/lambdag_pair_known_en}}
\end{minipage}\hfill
\begin{minipage}[t]{0.48\textwidth}
\textit{Questioned document (masked)}\\[3pt]
\begin{minipage}[t][\gerheight][t]{\textwidth}\input{texfiles/lambdag_pair_quest}\end{minipage}\\[4pt]
{\color{gray}\textit{Translation}\\[2pt]\itshape\input{texfiles/lambdag_pair_quest_en}}
\end{minipage}
\caption{Token-level $\lambda_G$ decomposition on \posNoise-masked transcripts of two documents by the same speaker. Content words are replaced by placeholders, while function words remain, and darker red marks a larger contribution to the decision. Omitted passages are indicated by [...]. The grey italic lines below each document are an English translation shown for readability only and carry no analysis of their own.}
\label{fig:lambdag}
\end{figure*}

\section{Conclusion}
\label{Conclusion}
We investigated AV for German spoken language using transcripts of video recordings. Our evaluation covered ten representative AV methods on three self-compiled corpora under multiple experimental settings, comparing their performance on both the original and \posNoise-masked transcripts.
%-------------------------------------------------------------------------- 
The traditional methods based on character- and grammar-level \ngram representations performed best, and no transformer-based method exceeded an accuracy of 0.750 under either condition. A plausible explanation is that \coav and \lambdaG rely on function words, word order and syntax, which automatic transcription and \posNoise preserve, whereas the transformer-based methods, pre-trained on written text and adapted from only 40 verification cases, are left with little beyond the topical cues removed by masking. This likely explains why \lambdaG was the only method to improve under masking across all three corpora. The token-level decomposition in Fig.~\ref{fig:lambdag} makes this concrete: in the case shown, the decision is carried by clause-initial connectives and function words rather than by domain vocabulary, which is exactly the kind of signal that survives topic masking. This decomposability is itself a practical advantage, as forensic settings require decisions that can be inspected rather than merely reported.
%-------------------------------------------------------------------------- 
This mirrors the findings of Aggazzotti et al. \cite{AggazzottiAAonSpeechTranscripts:2024}, who showed that the apparent success of AV on speech transcripts is partly driven by topical rather than stylistic cues and that controlling for topic reduces the performance of all models. The behaviour of \siamBERTFull illustrates the instability of the neural methods most clearly: masking raised its AUC on \CorpusFinance from 0.774 to 0.861 but lowered it on \CorpusDIY from 0.851 to 0.718, so the same method benefited from topic removal on one corpus and was harmed by it on another. Across all corpora and both conditions, the best non-neural method outperformed every transformer-based method.
%-------------------------------------------------------------------------- 
The error profiles under masking further separate the two families. Several neural and embedding-based methods combined high recall with low precision, accepting most same-author pairs at the cost of many false positives, whereas \coav on masked \CorpusDIY reached a precision of 0.96 at a recall of 0.80. For forensic use, where a false attribution and a missed attribution carry different costs, this difference between a cautious and a permissive verifier is as relevant as the headline accuracy.
%-------------------------------------------------------------------------- 
\styloSpeaker, the only method designed for speech transcripts, still stays behind. Its most discriminative features are filler words, backchannels and laughter annotations \cite{AggazzottiStyloSpeaker:2025}, none available here, and its many features on few training pairs make overfitting an equally plausible cause we cannot rule out. This points to a broader tension in our setup: the preprocessing that removes topical and disfluency cues in order to isolate style also removes the very features some speech-oriented methods depend on, so the ranking we report is conditional on these design choices.
%-------------------------------------------------------------------------
\section{Limitations} \label{Limitations}
%-------------------------------------------------------------------------- 
\subsection{Scope and Statistical Power}
Our corpora are small. The author-disjoint split assigns 20 speakers to training and 30 to testing, which yields only 40 verification cases per corpus for training and 60 for testing. The training set is too small to fit the transformer-based methods effectively, so their results characterize our setup rather than their attainable effectiveness. The test set has its own consequence, as a single verification case shifts accuracy by roughly 1.7 percentage points, so small differences between neighbouring methods fall within the resolution of the evaluation and should not be read as a ranking. We further report a single author-disjoint split and the median of three runs without confidence intervals, so only the larger performance gaps can be interpreted with confidence. A separate constraint concerns the transcription itself. Without manually produced reference transcripts we cannot quantify the word error rate of our automatic transcription, and we can therefore not separate errors introduced by the recognition system from genuine speaker characteristics. Finally, our findings are confined to German, to monologue speech and to three domains, so the observed behaviour of the methods may not carry over to other languages, registers or topics.
%-------------------------------------------------------------------------- 
\subsection{Ethical Considerations and Data Availability}
The speakers in our corpora are identifiable individuals who did not consent to the study, which is why we processed the material under the research exception and reduced the personal information it carries. All transcripts were anonymized by replacing personal and channel names with placeholders. This step relies on rule-based matching and named entity recognition and cannot guarantee complete removal of every identifying mention. No audio, video or video identifier is released. We note that topic masking removes content but not idiolect, and our own results show that a speaker can still be verified on masked text with an accuracy of up to 0.883 and an AUC of up to 0.907. Masked transcripts therefore remain pseudonymous rather than anonymous, and we release them only to researchers upon request and under an agreement that forbids re-identification. Speakers may request removal of their material at any time.
%-------------------------------------------------------------------------
%--------------------------------------------------------------------------
\section{Future Work} \label{FutureWork}
%-------------------------------------------------------------------------- 
Beyond the constraints noted above, our results suggest extending the verification task itself. Prior work on speech transcripts considered English telephone conversations, where each speaker's side is transcribed separately, and we considered videos deliberately restricted to a single speaker. In both cases each document corresponds to exactly one speaker. This assumption is granted by the recording setup rather than by the data itself, and it does not hold for unconstrained recordings. A natural generalisation is to recordings with several speakers, which turns the problem into a multi-speaker setting: rather than deciding whether two documents share an author, the task becomes deciding which of several known speakers, if any, is present in an unknown recording. A closely related reformulation is stylometric speaker segmentation, where the goal is not only whether a target speaker appears in an unknown video but also at which points, effectively a style-based counterpart to speaker diarisation on transcripts rather than on the acoustic signal. Such formulations would test whether the topic-agnostic features that succeeded in our pairwise setting remain discriminative when a document is no longer attributable to a single speaker. A further direction concerns the type of audio. We deliberately restricted our corpora to single-speaker monologue videos, but other genres, such as political speeches, broadcast interviews or debates, differ in register, preparation and speaker overlap, and would show whether the ranking of methods we observed is specific to prepared monologue or holds more broadly for transcribed speech.  final direction concerns language. Prior work on speech transcripts and our own evaluation are each confined to a single language, yet the methods that performed best rely on part-of-speech sequences and word order. It is therefore an open question whether the same grammatical signature is equally discriminative in morphologically rich languages such as Russian or Arabic, or in languages with different word-order regularities and without explicit word boundaries such as Chinese or Japanese. Establishing whether the ranking of methods we observed transfers across typologically diverse languages would show how far the present results extend beyond the single-language studies conducted so far.
%--------------------------------------------------------------------------
%------------------------------------------------ 

%\section*{References}
\newcommand{\BIBdecl}{\setlength{\itemsep}{0pt}}
\bibliographystyle{IEEEtran}
\bibliography{references}
\end{document}